# Extension of Rough Set Based on Positive Transitive Relation

Min Shu[1, 2, *], Wei Zhu[1, 2]

[1] Department of Applied Mathematics & Statistics, Stony Brook University, Stony Brook, NY, USA
[2] Center of Excellence in Wireless & Information Technology, Stony Brook University, Stony Brook, NY, USA

## Abstract

The application of rough set theory in incomplete information systems is a key problem in practice since missing values almost always occur in knowledge acquisition due to the error of data measuring, the limitation of data collection, or the limitation of data comprehension, etc. An incomplete information system is mainly processed by compressing the indiscernibility relation. The existing rough set extension models based on tolerance or symmetric similarity relations typically discard one relation among the reflexive, symmetric and transitive relations, especially the transitive relation. In order to overcome the limitations of the current rough set extension models, we define a new relation called the positive transitive relation and then propose a novel rough set extension model built upon which. The new model holds the merit of the existing rough set extension models while avoids their limitations of discarding transitivity or symmetry. In comparison to the existing extension models, the proposed model has a better performance in processing the incomplete information systems while substantially reducing the computational complexity, taking into account the relation of tolerance and similarity of positive transitivity, and supplementing the related theories in accordance to the intuitive classification of incomplete information. In summary, the positive transitive relation can improve current theoretical analysis of incomplete information systems and the newly proposed extension model is more suitable for processing incomplete information systems and has a broad application prospect.

*Corresponding author at: Department of Applied Mathematics & Statistics, Physics A149, Stony Brook University, Stony Brook, NY 11794, USA.
*E-mail address:* min.shu@stonybrook.edu (M. Shu), wei.zhu@stonybrook.edu (W. Zhu)

## 1. Introduction

The rough set theory developed by Pawlak [1] has become an important mathematical tool to deal with inexact and uncertain information after probability theory, fuzzy set theory and evidence theory. It serves as a component of hybrid solutions to machine learning and data mining problems, particularly for rule induction and feature selection. In terms of data analysis, rough set theory is advantageous in that any prior or additional information about the data is not required, such as the possibility values, probability distributions, basic probability assignments, or membership grades [2]. Rough set-based data analysis methods have found great successes in bioinformatics, economics and finance, medicine, multimedia, web and text mining, signal and image processing, software engineering, robotics, etc. The research and applications of rough set theory have attracted increasing attention in the recent years.

The rough set concept can be defined via the interior and closure topological operations approximating the upper and lower bounds of a set [1]. An information system can be viewed as a data table consisting of rows labeled by objects of interest, columns labeled by attributes, and table entries representing attribute values. A binary relation between two objects or more, known as an indiscernibility relation which is a central concept in rough set theory, can be derived from certain subsets of considered attributes in an information system. The indiscernibility relation is an equivalence relation where all identical objects of a set are considered as elementary [3]. Based on the indiscernibility relation, the fundamental concept of a rough set model: the approximation of lower and upper spaces of a set can be defined, which is the approximation of spaces for the formal classification of knowledge regarding the interest domain. The lower approximations generate the subset of objects that will definitely form part of an interest subset, whereas the subset generated by the upper approximation consists of objects that will possibly form part of an interest subset. Every subset defined through these two rough approximations is known as a rough set.

The classical rough set theory is based on complete information systems. A data table to be processed must be complete and all objects' values must be known. However, in practice, the incomplete information systems with missing values usually occur in knowledge acquisition due to errors in data measuring, the limitation of data acquisition or comprehension, etc. In order to process the incomplete information systems, two methods in rough set theory are usually used. The first is an indirect method called data reparation by transforming an incomplete information system to a complete information system using probabilistic and statistical technique [4]. The second is a direct method that extends the related concepts of the classical rough set theory to deal with incomplete information systems [5-10]. Existing rough set extension models based on the tolerance or symmetric similarity relation typically discards one relation among the reflexive, symmetric and transitive relations, especially the transitive relation.

In this paper, we first briefly summarize extensions to the classical rough set theory. Then, we propose a new relation called positive transitive relation and subsequently develop a novel extension of rough set model based on this new relation. The new model possesses the merit of the existing extensions of the classical rough set models while avoids their limitations of discarding transitivity or symmetry. It is thus more suitable for processing incomplete information systems.

## 2. Existing Rough Set Extension Models

The current extension models are mainly based on tolerance relation, similarity relation and limited tolerance relation.

**2.1** Tolerance relation are reflexive and symmetric, but completely non-transitive. It is defined as:

$$T=\{<x,y> \in U \times U| \ \forall a \in A,(a(x)=*) \vee ((a(y)=*) \vee ((a(x)=a(y))\}$$

**2.2** Similar relation are spontaneous and transitive, however it is not symmetry at all. It is defined as:

$$S=\{<x,y> \in U \times U| \ \forall a \in A,(a(x)=*) \vee ((a(x)=a(y))\}$$

**2.3** The limited tolerance relation is reflexive and symmetric, but not transitive. It is defined as:

$$L=\{<x,y> \in U \times U|a \in A(a(x)=a(y)=*) \vee ((P(x))P(y) \neq \phi \wedge a \in A((a(x) \neq *) \wedge (a(y) \neq *)$$
$$\rightarrow (a(x)=a(y))))\}$$, where $P(x)=\{a \in A|a(x) \neq *\}$

**2.4** Set-value-based analysis is another form of tolerance and similarity. The k-degree limit tolerance is extended based on limited tolerance.

The results from the above relations are limited since these relations do not have 1/3 of the equivalence properties. It is obviously not reasonable that two objects known to have equal attribute properties are not considered to be in the same class. For example, $x=(3,2,1,0)$ and $y=(3,2,1,3)$ are not considered to be in the same class on the basis of $z=(3,2,1,*)$ ). Therefore, it is more important to make reparation to the near-equivalence property.

## 3. Positive Transitive Relation

We shall examine a few examples below:

Example 1 {{3, 2, 0}, {3, 2, *}, {3, 2, 3}};
Example 2 {{1, *, 3}, {1, 3, *}, {1, *, 4}};
Example 3 {{3, 2, 0}, {3, *, 3}};
Example 4 {{3, 2, 0}, {3, *, 3}, {3, *, *}}.

The {3, 2, *} in Example 1 can provide an indistinguishable relation of the third position, so that a positive transitive relation can be constructed. In other words, {3, 2, 0}, {3, 2, 3} can be considered as a 2/3 indistinguishable relation in the presence of {3, 2, *} and can be classified into a restricted tolerance class.

Example 2 is a counterexample. It is not enough to be a positive transitive relation. In an incomplete information system, this transitive is irreparable and considered to be negative.

Because the existence of {1, 3, *} indicates that the third position can be constructed only if the second position is 3, that is, the degree of freedom of the second position is not provided. Thus, the set of {{1, *, 3}, {1, 3, *}, {1, 3, 4}} has the distinguishable relation, but the original example does not have the distinguishable relation.

Example 3 is not enough to make any distinguishable relation, but Example 4 works. It was previously thought that the set of {{3, 2, 0}, {3, 2, *}} constitutes a tolerance class, the set of {{3, 2, *}, {3, 2, 3}} constitutes a tolerance class, but the set of {{3, 2, 0}, {3, 2, 3}} does not constitute a tolerance class. In fact, {3, *, *} can constitute a bridge of positive transmission, making them constitute 1/3 of the indistinguishable relation. This is the reason behind our proposing the positive transitive relation.

Through the above examples, we can comprehend the basic concept of the positive transitive: under the premise that the values of the remaining three positions are constant, the uncertainty value at the same position can provide the change interval for the same position of the other two quantities, which constitutes positive transitive relation. The positive transitive relation must follow two conditions: First, for any set of 3 quantities, the corresponding position of the third quantity must be *. Second, the number or certainty of the non-variable position without * must be the same, and can no longer be * or the amount of uncertainty.

## 4. Rough Set Extension Model Based on Positive Transitive Relation

Under an incomplete information system:

$\mathrm{K}=(\mathrm{U},\mathrm{C}\cup\{\mathrm{d}\},\mathrm{V},\mathrm{f}),\mathrm{B}\subseteq C,\mathrm{x},\mathrm{y}\in\mathrm{U},\mathrm{P}_{\mathrm{B}}(\mathrm{x},\mathrm{y})=\{\mathrm{b}|\mathrm{b}\in\mathrm{B},\ \mathrm{b}(\mathrm{x})=\mathrm{b}(\mathrm{y})\neq *\}$
where $\mathrm{k}=|\mathrm{P}_{\mathrm{B}}(\mathrm{x},\mathrm{y})|/|\mathrm{B}|$ is equivalent to the degree of x and y on B $(0\leq k\leq 1)$.

Definition 1: A positive transitive relation is defined as:

$$\mathrm{M}=\{<\mathrm{x},\mathrm{y},\mathrm{z}>\in\mathrm{U}\times\mathrm{U}\times\mathrm{U}|\ (((\mathrm{x},\mathrm{y})\in\mathrm{T}\wedge(\mathrm{y},z)\in\mathrm{S})\vee((\mathrm{x},\mathrm{z})\in\mathrm{T}\wedge(\mathrm{z},\mathrm{y})\in\mathrm{S})\vee$$

$((\mathrm{z},\mathrm{x})\in\mathrm{T}\wedge(\mathrm{x},\mathrm{y})\in\mathrm{S})\vee\{<\mathrm{x},\mathrm{y}>\in\mathrm{U}\times\mathrm{U}|\ \forall\mathrm{a}\in\mathrm{A},\ (\mathrm{x},\mathrm{y})\in\mathrm{T}\}$ (T is the limit tolerance relation for k equivalence, and S is a similar relation.)

In the above relation, S has the relations of reflexivity, symmetry, and a certain non-losable positive transitivity. It can be simply seen as a tolerance relation plus a positive transitive relation.

Definition 2: The upper and lower approximation sets under the positive transitive relation are defined as:

$$\underline{\mathrm{M}}=\{\mathrm{X}\in\mathrm{U}|\mathrm{M}(\mathrm{x})\subseteq\mathrm{X}\}\quad \overline{\mathrm{M}}=\{\mathrm{X}\in\mathrm{U}|\mathrm{M}(\mathrm{x})\cap\mathrm{X}\neq\Phi\}$$

**Theorem 1:** In the incomplete information system $\mathrm{S}=(\mathrm{U},\mathrm{A},\mathrm{V},\mathrm{f})$, M is a positive transitive relation. When the intermediate transitive amount H and the E thus constituted positive transitive belong to the same class on the decision attribute, the positive transitive relation can reduce the complexity without affecting the classification result**.**

## 5. Performance Analysis of the Rough Set Extension Model Based on Positive Transitive Relation

Regarding the indiscernibility relation, the existing tolerance relations do not contain positive transitivity, resulting in increasing engineering computational complexity. However, the tolerance relation based on positive transitive relation can render it more realistic, significantly improve the reliability of theoretical analysis and maximize the close to equivalence relation, which in turn provides a basis for applying theoretical analysis of equivalence relations. The proposed extension model also adds complete symmetry to similar relations and retains the useful positive transitive relation, leading to more reliable results. Furthermore, some analysis based on set-values can also be supplemented. Therefore, the positive transitive relation can overcome the shortcomings of the current theoretical analysis of incomplete information systems.

The performance analysis of these extended rough set models is carried out using an actual incomplete information system in Table 1. Stefanowski and Tsoukiàs [6] used this incomplete information table to analyze the nature of tolerance relations, similar relations, and similar tolerance relations. For consistence, this information table is adopted to compare and analyze the nature of the positive transitive relation.

Table 1. Incomplete information table

| A | $c_1$ | $c_2$ | $c_3$ | $c_4$ | d |
|---|---|---|---|---|---|
| $a_1$ | 3 | 2 | 1 | 0 | Φ |
| $a_2$ | 2 | 3 | 2 | 0 | Φ |
| $a_3$ | 2 | 3 | 2 | 0 | Ψ |
| $a_4$ | * | 2 | * | 1 | Φ |
| $a_5$ | * | 2 | * | 1 | Ψ |
| $a_6$ | 2 | 3 | 2 | 1 | Ψ |
| $a_7$ | 3 | * | * | 3 | Φ |
| $a_8$ | * | 0 | 0 | * | Ψ |
| $a_9$ | 3 | 2 | 1 | 3 | Ψ |
| $a_{10}$ | 1 | * | * | * | Φ |
| $a_{11}$ | * | 2 | * | * | Ψ |
| $a_{12}$ | 3 | 2 | 1 | * | Φ |

The k-equivalence tolerance relation is firstly used for analysis. In order to compare with the limit tolerance at the same time, k = 0.25 is selected first. In the case of four conditional attributes, the two results are the same, as follows:

$I_C^L(a_1) = \{a_1, a_{11}, a_{12}\}$, $I_C^L(a_2) = \{a_2, a_3\}$, $I_C^L(a_3) = \{a_2, a_3\}$, $I_C^L(a_4) = \{a_4, a_5, a_{11}, a_{12}\}$,
$I_C^L(a_5) = \{a_4, a_5, a_{11}, a_{12}\}$, $I_C^L(a_6) = \{a_6\}$, $I_C^L(a_7) = \{a_7, a_9, a_{12}\}$, $I_C^L(a_8) = \{a_8\}$,
$I_C^L(a_9) = \{a_7, a_9, a_{11}, a_{12}\}$, $I_C^L(a_{10}) = \{a_{10}\}$, $I_C^L(a_{11}) = \{a_1, a_4, a_5, a_9, a_{11}, a_{12}\}$,
$I_C^L(a_{12}) = \{a_1, a_4, a_5, a_7, a_9, a_{11}, a_{12}\}$; $\Phi_C^L = \{a_{10}\}$, $\Psi_C^L = \{a_6, a_8\}$;
$\Phi_L^C = \{a_1, a_2, a_3, a_4, a_5, a_7, a_9, a_{10}, a_{11}, a_{12}\}$, $\Psi_L^C = \{a_1, a_2, a_3, a_4, a_5, a_6, a_7, a_8, a_9, a_{11}, a_{12}\}$

By applying the positive transitive relation to analyze the incomplete information table (Table 1), we shall obtain the following results:

$M_c^T(a_1) = \{a_1, a_4, a_5, a_9, a_{11}, a_{12}\}$, $M_c^T(a_2) = \{a_2, a_3\}$, $M_c^T(a_3) = \{a_2, a_3\}$,
$M_c^T(a_4) = \{a_1, a_4, a_5, a_9, a_{11}, a_{12}\}$, $M_c^T(a_5) = \{a_1, a_4, a_5, a_9, a_{11}, a_{12}\}$, $M_c^T(a_6) = \{a_6\}$,
$M_c^T(a_7) = \{a_1, a_7, a_9, a_{12}\}$, $M_c^T(a_8) = \{a_8\}$, $M_c^T(a_9) = \{a_1, a_4, a_5, a_7, a_9, a_{11}, a_{12}\}$,
$M_c^T(a_{10}) = \{a_{10}\}$, $M_c^T(a_{11}) = \{a_1, a_4, a_5, a_9, a_{11}, a_{12}\}$, $M_c^T(a_{12}) = \{a_1, a_4, a_5, a_7, a_9, a_{11}, a_{12}\}$;
$\phi_C^L = \{a_{10}\}$, $\psi_C^L = \{a_6, a_8\}$; $\phi_L^C = \{a_1, a_2, a_3, a_4, a_5, a_7, a_9, a_{10}, a_{11}, a_{12}\}$,
$\psi_L^C = \{a_1, a_2, a_3, a_4, a_5, a_6, a_7, a_8, a_9, a_{11}, a_{12}\}$.

Obviously, using positive transitive relation, the results of several categories that should be assigned to the same class fall into the same class. At the same time, the lower and upper approximation results are not greatly changed, and the positive transitive can make reparation to part of the classification loss. For example, the $a_1$ in $M_c^T(a_4)$ is compensated to the equivalence class of $a_9$ through $a_{12}$.

## 6. Application of extended model in knowledge reduction

Definition 5: Extended discernibility matrix elements: The discernibility matrix of decision table S is a matrix, and any one of which is:

$$a^*(x,y) = \begin{cases} \left(f(x,a) \neq f(y,a)\right) \wedge \left(f(x,a) \neq *\right) \\ \wedge \left(f(y,x) \neq *\right), \{f(x), f(y)\} \in U/ind(\overline{C}) \\ \phi, \{f(x), f(y)\} \in U/ind(\overline{C}) \end{cases}$$

In the decision table matrix reduction, the element w(x, y) must satisfy the following conditions:

$$\begin{aligned} & x \in pos_C(D) \text{ and } y \notin pos_C(D) \\ & \text{or } x \notin pos_C(D) \text{ and } y \in pos_C(D) \\ & \text{or } x, y \in pos_C(D) \text{ and } (x, y) \notin ind(D) \end{aligned}$$

**6.1** Knowledge reduction of decision tables by compatible relations

$$U/D = a_1, a_2, a_4, a_7, a_{10}, a_{12}; a_3, a_5, a_6, a_8, a_9, a_{11}$$
$$U/ind(C) = \{\{a_1, a_{12}\}, \{a_2, a_3\}, \{a_4, a_5, a_{11}\}, a_6, \{a_7, a_9\}, a_8, a_{10}\}$$
$$pos_C(D) = \{a_6, a_8, a_{10}\}$$

The resulting discernibility matrix is as follows:

$$\begin{array}{c} a_1 \\ a_2 \\ a_3 \\ a_4 \\ a_5 \\ a_6 \\ a_7 \\ a_8 \\ a_9 \\ a_{10} \\ a_{11} \\ a_{12} \end{array}
\left( \begin{array}{ccccccccccc}
 & & & & & & & & & & \\
1,2,3 & & & & & & & & & & \\
1,2,3 & \phi & & & & & & & & & \\
4 & 2,4 & 2,4 & & & & & & & & \\
4 & 2,4 & 2,4 & \phi & & & & \text{Symmetric} & & & \\
1,2,3,4 & 4 & 4 & 2 & 2 & & & & & & \\
4 & 1,4 & 1,4 & 4 & 4 & 1,4 & & & & & \\
2,3 & 2,3 & 2,3 & 2 & 2 & 2,3 \to \phi & \phi & & & & \\
4 & 1,2,3,4 & 1,2,3,4 & \phi & \phi & 1,2,3,4 & \phi & 2,3 & & & \\
1 & 1 & 1 & \phi & \phi & 1 & 1 & \phi & 1 & & \\
\phi & 2 & 2 & \phi & \phi & 2 & \phi & 2 & \phi & \phi & \\
\phi & 1,2,3 & 1,2,3 & \phi & \phi & 1,2,3 & \phi & 2,3 & \phi & 1 & \phi
\end{array} \right)$$

Note: Both 8 and 6 in (8, 6) are attached to $pos_C(D)$ and ind($D$), and do not satisfy the w(x,y) condition. Thus, the value is null.

The distinguishing function obtained from the above discernibility matrix:

$$\begin{aligned} \Delta^* &= (c_1 \vee c_2 \vee c_3) \wedge c_4 \wedge (c_2 \vee c_3) \wedge (c_1 \vee c_2 \vee c_3 \vee c_4) \\ &\quad \wedge c_1 \wedge c_2 \wedge (c_2 \vee c_4) \wedge (c_1 \vee c_4) \\ &= c_1 \wedge c_2 \wedge c_4 \end{aligned}$$

Therefore, this decision table has only one reduction $\{c_1, c_2, c_4\}$, no nuclear.

**6.2** The decision table is roughly divided by a positive transitive relation:

$$U/ind(\overline{C}) = \{\{a_1, a_4, a_5, a_7, a_9, a_{11}, a_{12}\}, \{a_2, a_3\}, a_6, a_8, a_{10}\}$$
$$pos_C(D) = \{a_6, a_8, a_{10}\}$$

The resulting discernibility matrix is as follows:

$$
\begin{array}{c}
a_1 \\ a_2 \\ a_3 \\ a_4 \\ a_5 \\ a_6 \\ a_7 \\ a_8 \\ a_9 \\ a_{10} \\ a_{11} \\ a_{12}
\end{array}
\left(
\begin{array}{cccccccccccc}
 & & & & & & & & & & & \\
2 & & & & & & & & & & & \\
2 & \phi & & & & & & & & & & \\
\phi & 2 & 2 & & & & & & \text{Symmetric} & & & \\
\phi & 2 & 2 & \phi & & & & & & & & \\
2 & 4 & 4 & 2 & 2 & & & & & & & \\
\phi & 1,4 & 1,4 & \phi & \phi & 1,4 & & & & & & \\
2 & 2,3 & 2,3 & 2 & 2 & \phi & \phi & & & & & \\
\phi & 2 & 2 & \phi & \phi & 2 & \phi & 2 & & & & \\
\phi & 1 & 1 & \phi & \phi & 1 & 1 & \phi & 1 & & & \\
\phi & 2 & 2 & \phi & \phi & 2 & \phi & 2 & \phi & \phi & & \\
\phi & 2 & 2 & \phi & \phi & 2 & \phi & 2 & \phi & 1 & \phi &
\end{array}
\right)
$$

The discernibility function obtained by discerning the matrix above is:

$$
\begin{aligned}
&\Delta^* = c_4 \wedge (c_2 \vee c_3) \wedge c_1 \wedge c_2 \wedge (c_1 \vee c_4) \\
&= c_1 \wedge c_2 \wedge c_4
\end{aligned}
$$

Therefore, the reduction of this decision table is the same as the above result $\{c_1, c_2, c_4\}$. Obviously, the storage space occupied by the discernibility matrix is much smaller than the previous one, which significantly reduces the computational complexity and is more generalizable.

## 7. Conclusion

The application of rough set theory in incomplete information systems is a key issue in practice because data are almost always incomplete in reality. In order to effectively apply the incomplete information systems and overcome the limitations of existing rough set extension models, we propose the positive transitive relation and develop a novel rough set extension model based on this relation. Through the performance analysis of the proposed rough set extension model, we find that the new model can substantially reduce the computational complexity, account for the relation of tolerance and similarity of positive transitivity, and supplement the related theories in accordance with the intuitive classification of incomplete information. Furthermore, the proposed extension model can significantly decrease the computational cost in knowledge reduction. Therefore, the proposed positive transitive relation is shown to improve the current theoretical analysis of incomplete information systems and the subsequent new extension model has a broad application prospect.